\documentclass[pmlr,twocolumn]{jmlr}
\usepackage{booktabs}
\usepackage[load-configurations=version-1]{siunitx}

\jmlrvolume{ML4H Extended Abstract Arxiv Index}
\jmlryear{2020}
\jmlrsubmitted{2020}
\jmlrpublished{}
\jmlrworkshop{Machine Learning for Health (ML4H) 2020}

\title[]{Bayesian recurrent state space model for rs-fMRI}

\author{%
   \Name{Arunesh Mittal\textsuperscript{\textdagger}} \Email{arunesh.mittal@columbia.edu}\\
   \Name{Scott Linderman\textsuperscript{\textsection}} \Email{scott.linderman@stanford.edu}\\
   \Name{John Paisley\textsuperscript{\textdagger}} \Email{jpaisley@columbia.edu}\\
   \Name{Paul Sajda\textsuperscript{\textdagger}} \Email{psajda@columbia.edu}\\
   \addr \textsuperscript{\textdagger}Columbia University, \textsuperscript{\textsection}Stanford University
}

\usepackage{multicol}
\usepackage{amsmath, amssymb}
\usepackage{dsfont}
\newcommand{\ts}{\textsuperscript}
\usepackage{tikz}
\newcommand*\circled[1]{\tikz[baseline=(char.base)]{
            \node[shape=circle,draw,inner sep=2pt] (char) {#1};}}
\usepackage{caption}
\usepackage{algorithm}

\newcommand{\bftab}{\fontseries{b}\selectfont}
\usepackage{siunitx}
\usepackage[multiple]{footmisc}
\SetKwComment{Comment}{$\triangleright$\ }{}

\begin{document}

\maketitle

\begin{abstract}
 We propose a hierarchical Bayesian recurrent state space model for modeling switching network connectivity in resting state fMRI data. Our model allows us to uncover shared network patterns across disease conditions. We evaluate our method on the ADNI2 dataset by inferring latent state patterns corresponding to altered neural circuits in individuals with Mild Cognitive Impairment (MCI). In addition to states shared across healthy and individuals with MCI, we discover latent states that are predominantly observed in individuals with MCI. Our model outperforms current state of the art deep learning method on ADNI2 dataset.
\end{abstract}

%\begin{keywords}
%List of keywords
%\end{keywords}

\section{Introduction}
\label{sec:intro}
Resting state fMRI (rs-fMRI) measures the intrinsic spontaneous activity across brain regions, in the absence of any sensory or cognitive stimulus, and has proven to be a useful tool in understanding the functional architecture of the brain. The functional network patterns estimated from this spontaneous brain activity have prognostic, diagnostic, as well as interventional utility. Alterations in functional network patterns have been observed in neuropathologies such as Alzheimer's, Depression, Schizophrenia, Autism and ADHD \citep{zhang2010disease, lee2013resting, fox2010clinical}. %Consequently, resting state network analysis has been proposed for identification and localization of functional networks during preoperative neurosurgical planning \citep{hacker2019resting}. 
In contrast to task-based fMRI, rs-fMRI improves accessibility of these methods to a much wider patient population, as it can be used in patients with motor or cognitive deficits, language barriers, infants, and aging patient populations, unable to perform directed tasks. 
\cite{vidaurre2017brain} showed that there are hierarchically organized stochastic recurring network patterns in rs-fMRI data in healthy subjects, which are both heritable and associated with specific cognitive traits. Several previous studies have applied standard state space models to rs-fMRI data \citep{suk2016state, zhang2019estimating, eavani2013unsupervised, ou2013modeling, wee2014group, liu2011monte}, however, these models have two limitations: 1) They do not model the shared connectivity patterns across healthy and disease groups 2) They do not account for the recurrent state patterns observed by \cite{vidaurre2017brain}. 

Building upon these prior works, we propose a state space model to uncover the spatio-temporal correlation patterns observed in rs-fMRI data. We model both class specific recurrent state transitions, as well as shared covariance structure across the classes. This allows modeling of the state switching behavior across different classes, as well as being able to delineate the network patterns shared across all classes, from the patterns dominant in only a particular class. %We propose an efficient inference procedure for our  model and evaluate our model on rs-fMRI data from ADNI2, a multi-site neuro-imaging study for progression of Alzheimer's disease.

\begin{figure*}[ht]
    % trim={<left> <lower> <right> <upper>}
    \includegraphics[width=1.\textwidth, trim={3.5cm 8.5cm 3cm 8.25cm},clip ]{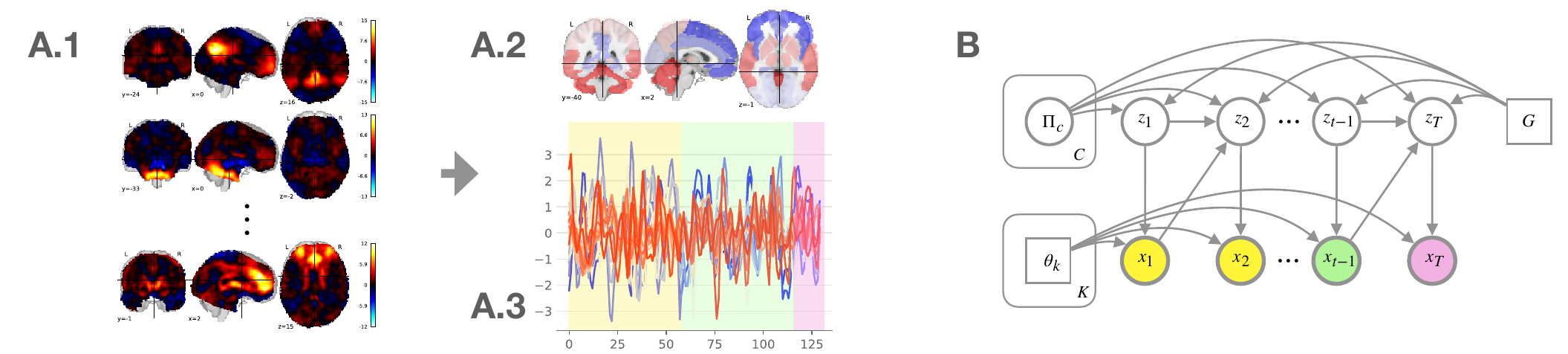}
    \captionsetup{labelformat=empty}
    \caption{
    \textbf{A}) 
    The rs-fMRI data for each individual (A.1) is aligned to a shared parcellation template (A.2) and the mean fMRI activity in each region is computed for all volumes in a sequence. This provides a lower dimensional temporal sequence $x_{1:T}$ for each individual (A.3). \textbf{B}) The probabilistic graphical model corresponding to the sequential generative process. The latent state sequence $z_t$ given $z_{t-1}, x_{t-1}$ for a class $c$ is modeled using an activity dependent transition matrix $\Psi_c^{(t)} \triangleq f(\Pi_c, G, x_{t-1})$. Given a state $z_{t} = k$ at time $t$, the rs-fMRI ROI mean and inter-ROI covariance is modeled as $\mathcal{N}(\mu_k, \Sigma_k)$. The spatial ROI correlations are modeled by $\Sigma_k$, while the temporal recurrence is modeled by $\Psi_c$, together these two processes capture the space and time dynamics in the data.
    }
    \label{fig:illus}
\end{figure*}

\section{Hierarchical recurrent state space model}
For a given sequence of fMRI activity $x_{1:T}$ across different regions of interest (ROI), we model the brain dynamics as a recurrent Markovian process where state $z_t \in \{1, \ldots, K \}$ at time $t$ depends on both the state at previous time step $z_{t-1}$, as well as the observations at the previous time step $x_{t-1} \in \mathbb{R}^D$ via a transition function $G: \mathbb{R}^D \rightarrow \mathbb{R}^K$. Given the latent state $z_t$ at a time $t$, the mean $\mu_{z_t}$ and the covariance $\Sigma_{z_t}$, model the mean and the covariance across all ROIs. Since changes in the BOLD signal in fMRI vary relatively slowly due to the hemodynamic response time, we impose a weak ``sticky" Dirichlet prior on the transition matrix to encourage slower state transitions. 

To model class specific differences in the brain across health and pathologic classes, we propose class conditional transition matrices $\{\Pi_c\}_{c \in \{1, \ldots, C\}}$. While the transition matrices are specific to each class, the observation parameters $\{\theta\}_{k=1}^{K} \triangleq \{\mu_k, \Sigma_k\}_{k=1}^{K}$ and transition function $G$, are shared across all classes. 

\begin{figure*}[t]
    \centering
    % trim={<left> <lower> <right> <upper>}
    \includegraphics[width=.9\textwidth, trim={1.cm 2.8cm 1cm 1.8cm},clip ]{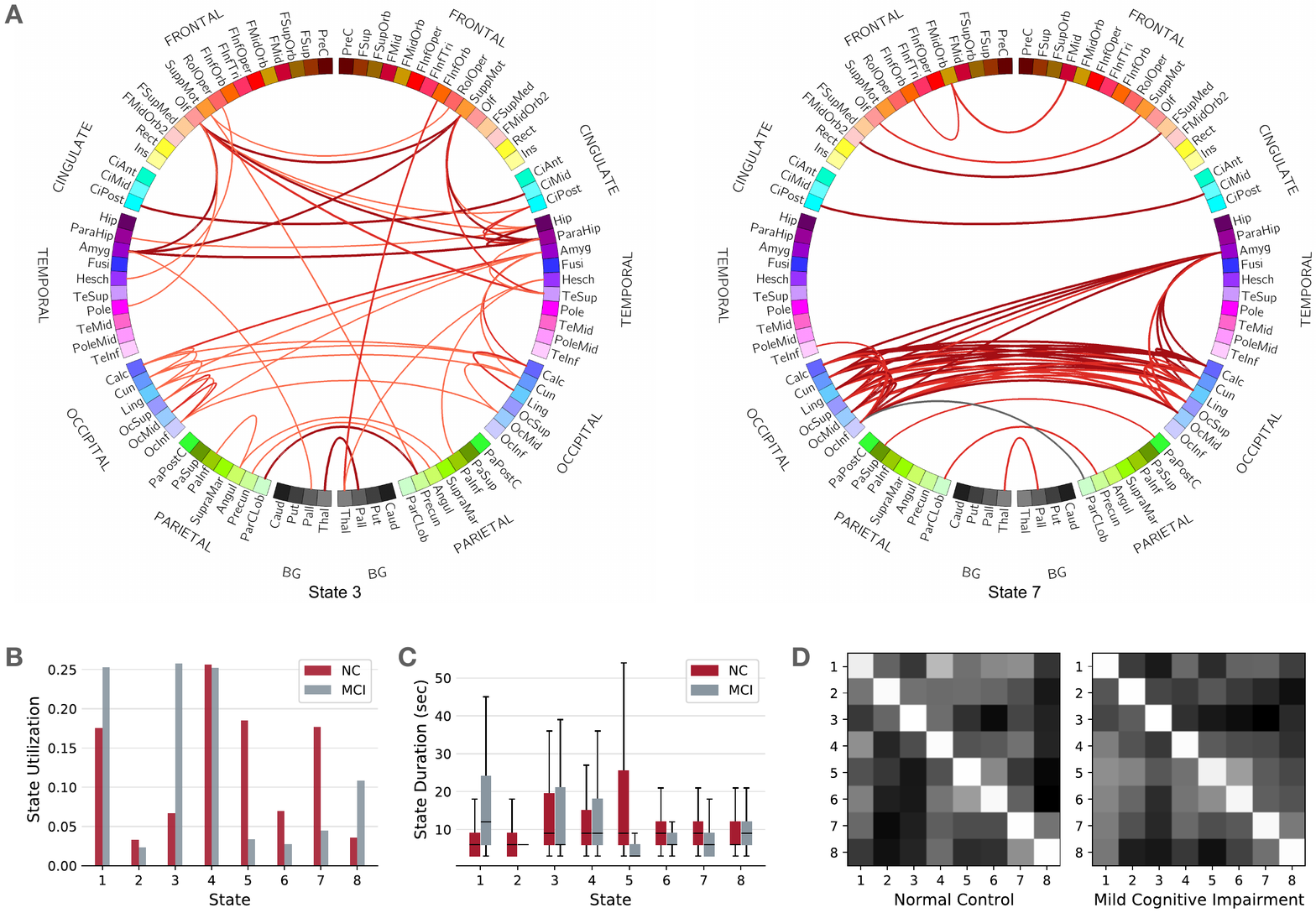}
    \captionsetup{labelformat=empty}
    \caption{
    \textbf{A}) Uncovered covariance across ROIs, for most discriminative latent states, $\Sigma_3$ and $\Sigma_7$, where for state $k$, a chord from region $i$ to $j$ corresponds to inter-ROI covariance $\Sigma_{k_{i,j}}$. The inner circle represents the anatomical regions for each ROI, the outer circle specifies the larger anatomical group the ROIs belongs to.   \textbf{B}) Relative state utilization by Normal Control vs. MCI individuals.
    \textbf{C}) State durations for Normal Control vs. MCI individuals.
    \textbf{D}) Class conditional transition matrices for NC and MCI.
    }
    \label{fig:results}
\end{figure*}

The class conditional transition matrices together with the shared observation parameters, allow us to infer the dominant class as well as shared latent states across the two classes. Additionally, this parameter tying allows us to compare the discovered latent structure across all disease classes (Fig \ref{fig:results}). This model structure encodes our belief that there are some states that are shared between the classes, and some states that are largely utilized by only one of the classes. The complete generative process is as follows:
\begin{gather*} %\label{eq1}
    y^{(n)} \sim \mathrm{Discrete}(p) \\
    [\Pi_{y^{(n)}}]_{k,:} \sim \mathrm{Dir}(\alpha 1_K + \kappa e_k) \\
    [\Psi_{y^{(n)}}]_{k,:}^{(t)} \triangleq \frac{[\Pi_{y^{(n)}}]_{k,:} \odot \exp(Gx_{t-1})^\top}{\mathcal{Z}_k} \\
    z_t^{(n)} \vert z_{t-1}^{(n)}  \sim [\Psi_{y^{(n)}}]_{k,:}^{(t)} \\
    x_t^{(n)}  \sim \mathcal{N}( \mu_{z^{(n)}_t}, \Sigma_{z^{(n)}_t})
\end{gather*} 

where $[\Pi_c]_{k,:} \in \Delta_{K-1}$ is the $k\ts{th}$ row of the $K \times K$ transition matrix $\Pi_c$ for class $c$, $\mathcal{Z}_k$ is normalizing constant for $k\ts{th}$ row of the recurrent transition matrix and $G \in \mathbb{R}^{K \times D}$ is the recurrent state perturbation matrix (See Appendix \ref{appendix:B}). 

\onecolumn
\begin{multicols}{2}

\section{Inference}
Our hierarchical model structure lends itself to a relatively simple modification of the standard message passing EM algorithm \citep{dempster1977maximum} for HMMs \citep{rabiner1989tutorial}. To maximize the joint likelihood of the data $p(x^{(n)}_{1:T}, z^{(n)}_{1:T}, \{\Pi_c\}_{c=1}^{C}, y^{(n)} \vert G, \{\theta\}_{k=1}^{K})$, we first marginalize out the latent states $z_{1:T}^{(n)}$ for each class $y^{(n)} \in \{1, \ldots, C\}$, and then proceed with the optimization M-step to update the shared parameters $\{\{\theta_k\}_{k=1}^{K}, G\}$ and the class specific parameters $\{\Pi_c\}_{c=1}^{C}$.

To marginalize the states in $\circled{\tiny{1}}$, for each class $y$, we only require the class conditional posterior marginals $q_c(z^{(n)}_t)  \triangleq p(z^{(n)}_t | \{\theta\}_{k=1}^{K}, \Pi_c, x_{1:T}^{(n)})$. For $\circled{\tiny{2}}$ we only require the class conditional two-slice marginal $q(z^{(n)}_t, z^{(n)}_{t-1}) \triangleq p(z^{(n)}_t, z^{(n)}_{t-1} | \{\theta\}_{k=1}^{K}, \Pi_c, x_{1:T}^{(n)})$. Hence, we can use the standard message passing algorithm to compute the class conditional posteriors for each class separately.

\begin{flalign*}
& \sum_{n,c}  \log p(x^{(n)}_{1:T}, \Pi, y^{(n)} \mid G, \theta) = &\\
\quad &  \sum_{n,c}  \underbrace{\mathbb{E}_{q_c(z)}\left[ \log p (x^{(n)}_t | z^{(n)}_t, \theta)^{\mathds{1}[y^{(n)} = c] }\right]}_{\circled{\tiny{1}}} \\
&+   \underbrace{\mathbb{E}_{q_c(z)} \left[ \log p(z^{(n)}_t | z^{(n)}_{t-1}, G, \Pi_c)^{\mathds{1}[y^{(n)} = c]} \right]}_{\circled{\tiny{2}}} \\ 
&+ \underbrace{\phantom{\Big[}\log p(\Pi_c)}_{\circled{\tiny{3}}} \nonumber
\end{flalign*}

\end{multicols}
\begin{table*}[h]
  \caption{Comparison with Deep-Autoencoder-HMM on ADNI2 dataset}
  \label{results-table}
  \centering
  \begin{tabular}{cccccc}
    \toprule
    Methods     & Accuracy     & Sensitivity  & Specificity & Pos. Pred. Val. & Neg. Pred. Val. \\
    \midrule
    Suk et al.  & 72.58 & 70.59 & 75 & \bftab 77.42 & 67.84 \\
    Ours                 & \bftab 79.17 & \bftab 83.33 & 75 & 76.92 &  \bftab81.82\\
    \bottomrule
  \end{tabular}
\end{table*}
\begin{multicols}{2}
   After marginalization, we then use gradient methods to update the the parameters $\{\{\theta\}_{k=1}^{K}, G, \{\Pi_c\}_{c=1}^{C}\}$, to maximize the complete data log-likelihood $\sum_{n,c}  \mathbb{E}_{q_c(z^{(n)})}\left[ \log p(x^{(n)}_{1:T}, z^{(n)}_{1:T}, \Pi, y^{(n)} \mid G, \theta) \right]$  (See Appendix \ref{appendix:A} and \ref{appendix:B}  for algorithm). 

\section{Experiments}
We constructed our dataset by extracting all rs-fMRI images from the ADNI2 \footnote{http://adni.loni.usc.edu/} dataset. We matched the age and gender between the two groups: subjects diagnosed with Mild Cognitive Impairment (Mean age: $71.72 \pm 2.82$) and normal control subjects (Mean age: $71.65 \pm 2.90$). This resulted in a dataset of $116$ subjects, with sequence length $T=130$. Following the pre-processing steps used by \cite{challis2015gaussian} (see: Fig. \ref{fig:illus}, Appendix \ref{appendix:D}), each brain volume was parcellated into physiologically meaningful pre-defined regions of interest by warping and masking each subjects scan using the AAL atlas by \cite{tzourio2002automated}. We then computed the mean time course for each region by computing the mean region of interest activity for each time point in a sequence. For training and test, we split up our dataset into age matched train ($N=92$) and test set ($N=24$). We chose hyper-parameters by 5-fold cross-validation on the train dataset. For the presented results we used $\{K=8, \alpha=.5, \kappa=100\}$. We report results on the heldout set in Table \ref{results-table}, where we predict the disease class using the Bayes classifier with uniform prior probability over classes: $\hat{y} = \underset{y}{\arg\max} \ \mathbb{E}_{q_y(z^{(n)})}\left[ \log p(x^{(n)}, z^{(n)}, \Pi, y \vert G, \theta) \right]$.

\section{Conclusion}
Our method allows us to model temporal recurrent dynamics, which are influenced by class specific local parameters $\{\Pi_c\}_{c=1}^{C}$ and class agnostic global parameter $G$. The hierarchical structure then lets us compare the network patterns associated with each of the disease classes. We compare the uncovered correlations associated with the most prevalent states in MCI and NC individuals. We observe stronger inter-hemispheric cortical connections in healthy individuals. In addition, we observe stronger inter-hemispheric connectivity across occipital regions in healthy subjects. These findings have been independently reported by previous voxel-mirrored homotopic connectivity analysis in individuals with MCI \citep{wang2015interhemispheric}. In future work, we plan to evaluate our method on additional rs-fMRI datasets for other neuropathologies. To improve the model capacity, we will explore a more flexible non-linear likelihood model. Lastly, given the high dimensionality of the voxel data, we also plan to investigate incorporating additional prior structure on the covariance matrices, reflective of the anatomical connectivity constraints. In addition to the phenotype matching approach proposed in this work, we hope to extend this approach to uncover latent phenotypes \citep{chen2020probabilistic} from rs-fMRI data. 

%\acks{Acknowledgements go here.}

\newpage
\bibliography{ref}
\end{multicols}
\newpage
\onecolumn
\appendix
\section{Full EM algorithm} 
\label{appendix:A}
\definecolor{dimgray}{rgb}{0.41, 0.41, 0.41}

\begin{algorithm}
  \DontPrintSemicolon
  \caption{EM for hierarchical recurrent state space model}\label{euclid}
    Initialize $\{\theta_k\}_{k=1}^{K} \triangleq \{\mu_k, \Sigma_k\}_{k=1}^{K}$ \Comment*[r]{\textcolor{dimgray}{Initialize with empirical Bayes GMM}} 
    \BlankLine
    Random Initialization $\{\{\Pi_c\}_{c \in \{1, \ldots, C\}}, G\}$ \;
    \BlankLine
    \BlankLine
    \For{$m \in \{1, \ldots, M\}$}{
        \Comment*[l]{\textcolor{dimgray}{M iterations}}
        \For{$n \in \{1, \ldots, N\}$}{\Comment*[l]{\textcolor{dimgray}{N ROI sequences}}
            \For{$c \in \{1, \ldots, C\}$}{\Comment*[l]{\textcolor{dimgray}{Forwards Backwards for each class}}
                $\{q_c(z^{(n)}_t), q_c(z^{(n)}_t, z^{(n)}_{t-1}) \} \gets \mathrm{ForwardsBackwards}(x_{1:T}^{(n)}, \{\theta_k\}_{k=1}^{K}, \Pi_c)$
            }
        }
        \BlankLine
        \BlankLine
        \For{$k \in \{1, \ldots, K\}$}{\Comment*[l]{\textcolor{dimgray}{Analytic update state parameters}}
            $\{\mu_k, \Sigma_k\} \gets \underset{\{\mu_k, \Sigma_k\}}{\mathrm{argmax}}\sum_y \sum_n \mathbb{E}_{q_c(z^{(n)}_{1:T})}\left[ \mathds{1}[y^{(n)} = c] \log p (x^{(n)}_t | z^{(n)}_t, \{\theta_k\}_{k=1}^{K})\right]$ %\Comment{Analytical M-step for Gauss. likelihood}
        }
        \BlankLine
        \BlankLine
        \For{$l \in \{1, \ldots, L\}$}{\Comment*[l]{\textcolor{dimgray}{L iterations}}
        \Comment*[l]{\textcolor{dimgray}{M-step for transition function}}
            $\widetilde{G} \gets \widetilde{G} + \eta \nabla_{\widetilde{G}} \sum_y \mathbb{E}_{q_c(z^{(n)}_{1:T})} \left[\log p(z^{(n)}_t | z^{(n)}_{t-1}, e^{(\widetilde{G})}, \Pi_c) \right]$ 
            \BlankLine
            \BlankLine
            \For{$c \in \{1, \ldots, C\}$}{\Comment*[l]{\textcolor{dimgray}{M-step for transition matrices}}
            $\widetilde{\Pi}_c \gets \widetilde{\Pi}_c + \eta \nabla_{\widetilde{\Pi}_c} \sum_c  \bigg(\mathds{1}[y^{(n)} = c] \mathbb{E}_{q_c(z^{(y)}_{1:T})} \left[ \log p(z^{(y)}_t | z^{(y)}_{t-1}, \log(\widetilde{G}), e^{(\widetilde{\Pi}_c)}) \right]   + \log p(e^{(\widetilde{\Pi}_c)}) \bigg)$
            }
        }
    }
\end{algorithm}

\section{Generative Model}
\label{appendix:B}
\subsection{Hierarchical recurrent Markov model }
\begin{flalign}
    &\prod_n p\left(x^{(n)}_{1:T},  y^{(n)}, \{\Pi_c\}_{c=1}^{C} \vert  G,
    \{\theta_k\}_{k=1}^K \right)  & \\
    &=  \prod_c \prod_n  \prod_t \left(p (x^{(n)}_t \vert z^{(n)}_{t},
             \{\theta_k\}_{k=1}^K)
          p(z^{(n)}_{t} \vert z^{(n)}_{t-1}, G, x^{(n)}_{t-1},
            \Pi_c)
          p(\Pi_c) p(y^{(n)}) \right)^{\mathds{1}[y^{(n)} = c]} & \\
    &=  \prod_c \prod_n  \prod_t  \prod_{i,j}  p (x^{(n)}_t 
           \vert \theta_{z^{(n)}_{t}})^{\mathds{1}[z^{(n)}_{t} = i]}
         \\
         &\phantom{{}=\prod_c \prod_n  \prod_t \prod_k } 
          \Big(p(z^{(n)}_{t} \vert z^{(n)}_{t-1}, G, x^{(n)}_{t-1},
            \Pi_c)^{\mathds{1}[z^{(n)}_{t} = i, z^{(n)}_{t-1} = j]}
          p(\Pi_c)  p(y^{(n)}) \Big)^{\mathds{1}[y^{(n)} = c]} &
\end{flalign}
\subsection{Recurrent transition model with sticky transition prior}

\begin{flalign}
    &p(z^{(n)}_{t} \vert z^{(n)}_{t-1} = k, G, x^{(n)}_{t-1}, \Pi_c)p(\Pi_c) &&\\
    & = \mathrm{Discrete}\left( z^{(n)}_{t}  \vert \frac{[\Pi_c]_{k,:} \odot \exp(Gx_{t-1})^\top}{\mathcal{Z}_k}\right)\mathrm{Dir}([\Pi_c]_{k,:} \vert \alpha 1_K + \kappa e_k) &&
\end{flalign}
Where $\mathcal{Z}_k$ is normalizing constant for $k\ts{th}$ row of the recurrent transition matrix and $G \in \mathbb{R}^{K \times D}$ is the recurrent state perturbation matrix.

\section{Inference}
\label{appendix:C}
\subsection{Expected complete data log-likelihood}

\begin{flalign}
    &\sum_n \log  \ p(x_{1:T}^{(n)} , \{\Pi_c\}_{c=1}^{C} \vert \{\theta_k\}_{k=1}^K, G) & \nonumber\\
    = &\sum_n  \mathbb{E}_{q(z^{(n)})}  \left [ \log  p(x_{1:T}^{(n)} , \{\Pi_c\}_{c=1}^{C} \vert \{\theta_k\}_{k=1}^K, G) \right ] \\
    = &\sum_c \sum_{\{n:y^{(n)}=c\}}  \sum_t   \sum_i \Bigg( q_c(z^{(n)}_{t} = i) 
      \log  p  (x^{(n)}_t     
      \vert \theta_{z^{(n)}_{t}})  \ + \\ 
    &\phantom{{}= \sum_c \sum_{\{n:y^{(n)}=c\}}  \sum_t }
      \sum_{j} q_c(z^{(n)}_{t}=i, z^{(n)}_{t-1}=j)  \log p(z^{(n)}_{t} \vert z^{(n)}_{t-1}, G, x^{(n)}_{t-1}, \Pi_c)\nonumber +
      \log p(\Pi_c) \Bigg)\\
    %&\sum_c \sum_{\{n:y^{(n)}=c\}}  \sum_t    & \nonumber
\end{flalign}

Where class conditional posterior $q_c(z_t^{(n)})$ and $q_c(z^{(n)}_{t}, z^{(n)}_{t-1})$ can be computed using the standard Baum-Welch algorithm.

\subsection{Gradient update for \texorpdfstring{$\Pi_c$}{Pi}}
$\Pi_c \triangleq \exp (\widetilde{\Pi}_c)$
\begin{flalign}
    \nabla_{\widetilde{\Pi}_c}&\sum_n \log p(x_{1:T}^{(n)};  \{\theta_k\}_{k=1}^K, G, \widetilde{\Pi}_c) & \\
    = & \sum_{\{n:y^{(n)}=c\}}  \sum_t  \sum_{i,j} \nabla_{\widetilde{\Pi}_c} q_c(z^{(n)}_{t}=i, z^{(n)}_{t-1}=j)  \log p(z^{(n)}_{t} \vert z^{(n)}_{t-1}, G, x^{(n)}_{t-1}, \widetilde{\Pi}_c)\nonumber 
\end{flalign}

\begin{flalign}
    \Pi_c \gets \exp \left( \widetilde{\Pi}_c + \rho \sum_{\{n:y^{(n)}=c\}}  \sum_t  \sum_{i,j} \nabla_{\widetilde{\Pi}_c} q_c(z^{(n)}_{t}=i, z^{(n)}_{t-1}=j)  \log p(z^{(n)}_{t} \vert z^{(n)}_{t-1}, G, x^{(n)}_{t-1}, \widetilde{\Pi}_c)  \right)
\end{flalign}

\section{Preprocessing}
\label{appendix:D}
We smoothed the 3D volumes with a $\SI{8}{\milli\meter\cubed}$ FWHM 3D Gaussian kernel, followed by filtering the data with a band-pass filter (0.01 to 0.08 Hz). We discarded the first 10 volumes of each sequence to avoid saturation effects. Each brain volume was parcellated into physiologically meaningful pre-defined regions of interest by warping and masking each subjects scan using the AAL atlas by \cite{tzourio2002automated}.  To reduce the dimensionality of the data, we discard atlas regions (90 - 116 inclusive), corresponding to the Cerebellum.

\end{document}